\documentclass[]{llncs}
\usepackage{graphicx}
\usepackage{comment}
\usepackage{amsmath,amssymb} % define this before the line numbering.
\usepackage{color}
\usepackage{hyperref}
\usepackage{url}
\usepackage{wrapfig}
\usepackage{subfigure}
\usepackage{cite}
\usepackage{caption}
\usepackage{makecell}
\usepackage{pifont}
\usepackage{enumitem}
\newcommand{\cmark}{\ding{51}}
\newcommand{\xmark}{\ding{55}}

\usepackage[width=122mm,left=12mm,paperwidth=146mm,height=193mm,top=12mm,paperheight=217mm]{geometry}

\begin{document}
% \linenumbers
\pagestyle{headings}
\mainmatter
\def\ECCVSubNumber{1196}  % Insert your submission number here

\title{MutualNet: Adaptive ConvNet via Mutual Learning from Network Width and Resolution} % Replace with your title
% Width-Resolution Mutual Learning for Improving Universal Accuracy-Efficiency Tradeoffs; Deep Width-Resolution Mutual Learning

%\begin{comment}
% \titlerunning{ECCV-20 submission ID \ECCVSubNumber} 
% \authorrunning{ECCV-20 submission ID \ECCVSubNumber} 
% \author{Anonymous ECCV submission}
% \institute{Paper ID \ECCVSubNumber}
%\end{comment}
%******************

% \begin{comment}
\titlerunning{Abbreviated paper title}
\author{Taojiannan Yang\inst{1} \and Sijie Zhu\inst{1}
 \and Chen Chen\inst{1} \and Shen Yan\inst{2} \and Mi Zhang\inst{2} \and Andrew Willis\inst{1}
}

\authorrunning{F. Author et al.}
\institute{University of North Carolina at Charlotte \\ \email{\{tyang30,szhu3,chen.chen,arwillis\}@uncc.edu} \and
Michigan State University
\\ \email{\{yanshen6,mizhang\}@msu.edu}}
% \end{comment}
%******************
\maketitle

\begin{abstract}
 We propose the width-resolution mutual learning method (MutualNet) to train a network that is executable at dynamic resource constraints to achieve adaptive accuracy-efficiency trade-offs at runtime. 
 %It takes different input scales and shares the representations across different network widths\footnote{Number of channels in a layer.}, which enables each sub-network to capture multi-scale representations.
 %The proposed framework is simple yet effective. 
 Our method trains a cohort of sub-networks with different widths\footnote{Number of channels in a layer.} using different input resolutions to mutually learn multi-scale representations for each sub-network.
 It achieves consistently better ImageNet top-1 accuracy over the state-of-the-art adaptive network US-Net \cite{uslimnet} under different computation constraints, and outperforms the best compound scaled MobileNet in EfficientNet \cite{efficientnet} by 1.5\%. The superiority of our method is also validated on COCO object detection and instance segmentation as well as transfer learning. Surprisingly, the training strategy of MutualNet can also boost the performance of a single network, which substantially outperforms the powerful AutoAugmentation \cite{cubuk2019autoaugment} in both efficiency (GPU search hours: 15000 vs. 0) and accuracy (ImageNet: 77.6\% vs. 78.6\%). Code is available at \url{https://github.com/taoyang1122/MutualNet}
\end{abstract}

\section{Introduction}

\begin{wrapfigure}{r}{0.43\textwidth}
\centering
    %\vspace{-0.75cm}
    \vspace{-\intextsep}
    \includegraphics[width=0.43\textwidth]{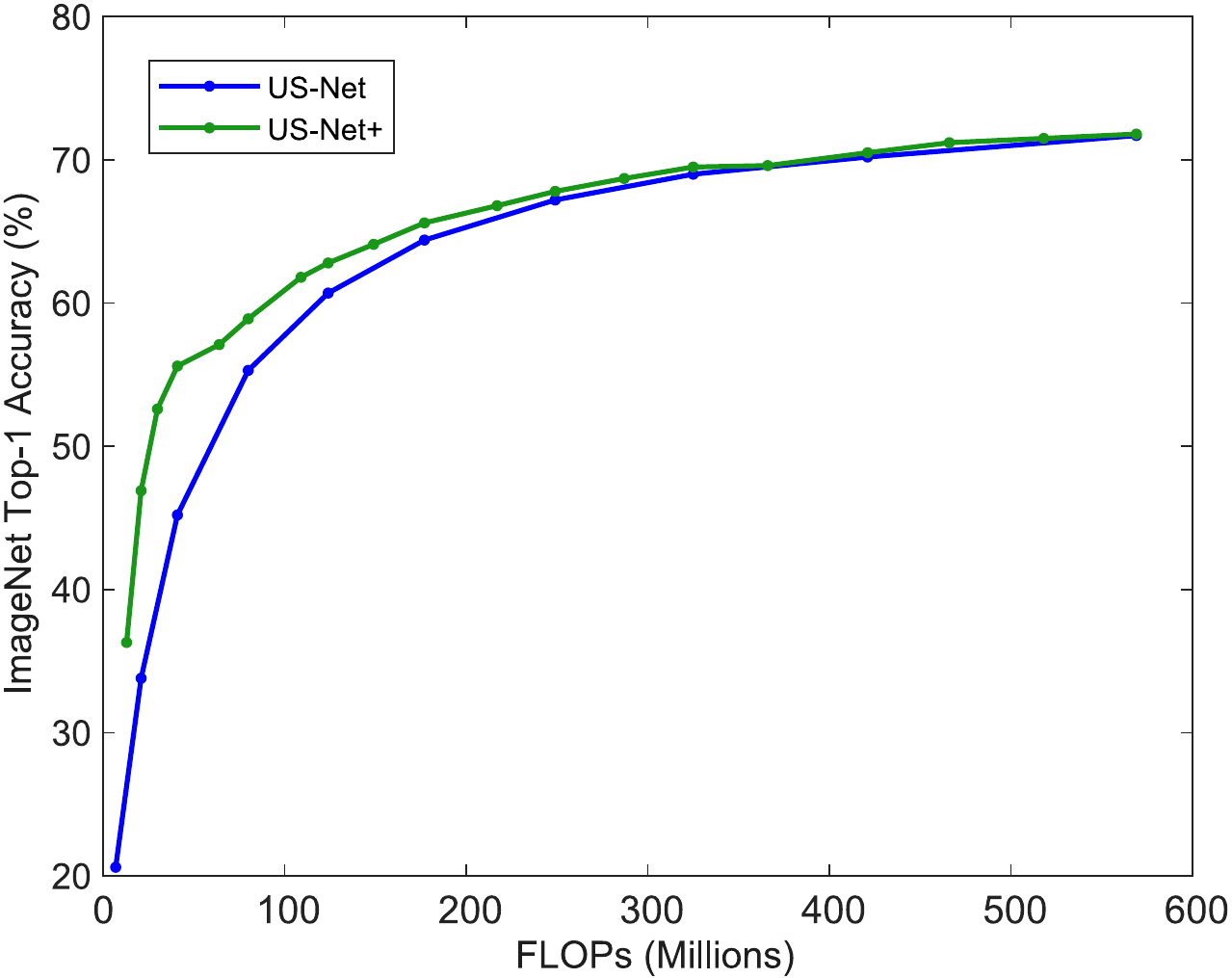}
    \vspace{-0.5cm}
    \caption{Accuracy-FLOPs curves of US-Net+ and US-Net.} %US-Net+ means simply applying different resolutions to US-Net during testing.
    \vspace{-0.7cm}
\end{wrapfigure}
\label{figure1}
%Deep neural networks have triumphed over various perception tasks such as image classification \cite{resnet, densenet, vgg}, object detection \cite{fasterrcnn, yolo} and semantic segmentation \cite{deeplab}. 
Deep neural networks have triumphed over various perception tasks.
However, deep networks usually require large computational resources, making them hard to deploy on mobile devices and embedded systems. This motivates research in reducing the redundancy in deep neural networks by designing efficient convolutional blocks \cite{howard2017mobilenets, zhang2018shufflenet, mobilenetv2, shufflenetv2} or pruning unimportant network connections \cite{strprune, structuredprune2, structured3}. However, these works ignore the fact that the computational cost is determined by both the network scale and input scale. Only focusing on reducing network scale cannot achieve the optimal accuracy-efficiency trade-off. EfficientNet \cite{efficientnet} has acknowledged the importance of balancing among network depth, width and resolution. 
%But it considers network scale and input scale as independent factors. 
But it considers network scale and input scale separately.
The authors conduct grid search over different configurations and choose the best-performed one, while we argue that \textit{network scale and input scale should be considered jointly in learning} to take full advantage of the information embedded in different configurations. 

Another issue that prevents deep networks from practical deployment is that the resource budget (e.g., battery condition) varies in real-world applications, while traditional networks are only able to run at a specific constraint (e.g., FLOP). To address this issue, SlimNets \cite{slimnet, uslimnet} are proposed to train a single model to meet the varying resource budget at runtime. They only reduce the network width to meet lower resource budgets. As a result, the model performance drops dramatically as computational resource goes down. Here, we provide a concrete example to show the importance of balancing between input resolution and network width for achieving better accuracy-efficiency trade-offs. Specifically, to meet a dynamic resource constraint from 13 to 569 MFLOPs on MobileNet v1 backbone, US-Net \cite{uslimnet} needs a network width range of [0.05,1.0]$\times$ given a 224$\times$224 input, while this constraint can also be met via a network width of [0.25,1.0]$\times$ by adjusting the input resolution from \{224, 192, 160, 128\} during test time. We denote the latter model as US-Net+. As shown in Fig. \ref{figure1}, simply combining different resolutions with network widths \textit{during inference} can achieve a better accuracy-efficiency trade-off than US-Net without additional efforts.
\begin{table}[t]
\caption{Comparison between our framework and previous works.}
\begin{center}
\footnotesize
\begin{tabular}{c|c|c|c|c}
    \hline
     Model& Adaptive & \textcolor{red}{N}etwork \textcolor{red}{S}cale & \textcolor{blue}{I}nput \textcolor{blue}{S}cale  & Mutual Learning (\textcolor{red}{NS}\&\textcolor{blue}{IS})   \\
     \hline
     MobileNet \cite{howard2017mobilenets, mobilenetv2}& \xmark& \cmark& \xmark& \xmark \\
     ShuffleNet \cite{zhang2018shufflenet, shufflenetv2}& \xmark& \cmark& \xmark& \xmark \\
     EfficientNet \cite{efficientnet}& \xmark& \cmark& \cmark& \xmark \\
     US-Net \cite{uslimnet}& \cmark& \cmark& \xmark& \xmark \\
     MutualNet (Ours) & \cmark& \cmark& \cmark& \cmark\\
     \hline
\end{tabular}
\end{center}
\label{comp}
\end{table}

\begin{figure}[t]
    \centering
    \includegraphics[width=0.9\linewidth]{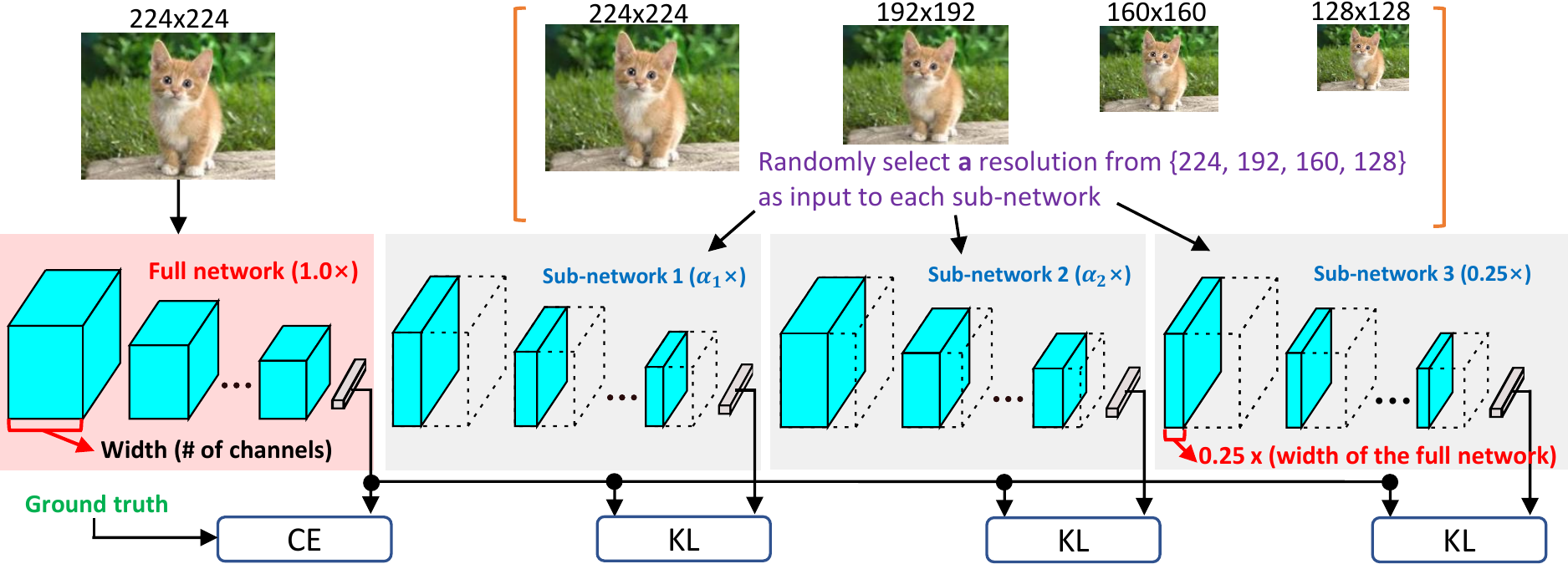}
    \caption{The training process of our proposed MutualNet. The network width range is [0.25, 1.0]$\times$, input resolution is chosen from \{224, 192, 160, 128\}. This can achieve a computation range of [13, 569] MFLOPs on MobileNet v1 backbone. We follow the \textit{sandwich rule} \cite{uslimnet} to sample 4 networks, i.e., upper-bound full width network ($1.0\times$), lower-bound width network ($0.25\times$), and \textbf{two random width ratios} $\alpha_1, \alpha_2 \in (0.25, 1)$. For the full-network, we constantly choose 224$\times$224 resolution. For the other three sub-networks, we randomly select its input resolution. The full-network is optimized with the ground-truth label. Sub-networks are optimized with the prediction of the full-network. Weights are shared among different networks to facilitate mutual learning. CE: Cross Entropy loss. KL: Kullback–Leibler Divergence loss.}
    \label{slimnet}
\end{figure}

Inspired by the observations above, we propose a \textit{mutual learning} scheme which incorporates both network width and input resolution into a unified learning framework. 
%Second, lower input resolution can reduce computations and does not necessarily harm the performance. \cite{chin2019adascale} points out that lower image resolution may produce better detection accuracy by reducing focus on redundant details. Third, different resolutions contain different information \cite{octave}. Lower resolution images may contain more global structures while higher resolution ones may encapsulate more fine-grained patterns. We can learn multi-scale representations from different resolutions.
As depicted in Fig.~\ref{slimnet}, our framework feeds different sub-networks with different input resolutions. Since sub-networks share weights with each other, each sub-network can learn the knowledge shared by other sub-networks, which enables them to capture \textit{multi-scale representations from both network scale and input resolution}.  %In this paper, we address these issues in a unified mutual learning framework. A comparison between our framework and previous works is in Table \ref{comp}.
Table \ref{comp} provides a comparison between our framework and previous works. In
summary, we make the following contributions:

\begin{itemize}[leftmargin=*]
   \item We highlight the importance of input resolution for efficient network design. Previous works either ignore it or treat it independently from network structure. In contrast, we embed network width and input resolution in a unified mutual learning framework to learn a deep neural network (MutualNet) that can achieve adaptive accuracy-efficiency trade-offs at runtime. 
   \item We carry out extensive experiments to demonstrate the effectiveness of our MutualNet. It significantly outperforms independently-trained networks and US-Net on various network structures, datasets and tasks under different constraints. To the best of our knowledge, we are the first to benchmark arbitrary-constraint adaptive networks on object detection and instance segmentation.
   \item We conduct comprehensive ablation studies to analyze the proposed mutual learning scheme. We further demonstrate that our framework is promising to serve as a plug-and-play strategy to boost the performance of a single network, which substantially outperforms the popular performance-boosting methods, e.g., data augmentations \cite{cubuk2019autoaugment, fastautoaug, cutout}, SENet \cite{senet} and knowledge distillation \cite{kdimprovement}.
   \item The proposed framework is a general training scheme and model-agnostic. It can be applied to any networks without making any adjustments to the structure. This makes it compatible with other state-of-the-art techniques (e.g., Neural Architecture Search (NAS) \cite{mobilenetv3, mnasnet}, AutoAugmentation \cite{cubuk2019autoaugment, fastautoaug}).
\end{itemize}

\section{Related Work}

{\bf Light-weight Network.} %In recent years, researchers are paying more attention to light-weight networks design. 
There has recently been a flurry of interest in designing light-weight networks. MobileNet~\cite{howard2017mobilenets} factorizes the standard $3\times3$ convolution into a $3\times3$ depthwise convolution and a $1\times1$ pointwise convolution which reduce computation cost by several times. ShuffleNet~\cite{zhang2018shufflenet} separates the $1\times1$ convolution into group convolutions to further boost computation efficiency.
%reduce the cost and proposes the Shuffle operation to help information flow between different groups. 
MobileNet v2~\cite{mobilenetv2} proposes the inverted residual and linear block for low-complexity networks. ShiftNet~\cite{wu2018shift} introduces a zero-flop shift operation to reduce computation cost. Most recent works \cite{wu2019fbnet, mnasnet, mobilenetv3} also apply neural architecture search methods to search efficient networks. However, none of them considers the varying resource constraint during runtime in real-world applications. \textit{To meet different resource budgets, these methods need to deploy several models and switch among them, which is not scalable}.

{\noindent\bf Adaptive Neural Networks.} To meet the dynamic constraints in real-world applications, MSDNet~\cite{multidense} proposes a multi-scale and coarse to fine densenet framework. It has multiple classifiers and can make early predictions to meet varying resource demands. NestedNet~\cite{kim2018nestednet} uses a nested sparse network which 
%exploited an n-in-1 nested structure. The nested sparse network 
consists of multiple levels
%and higher level networks share the parameters with lower level networks 
to enable nested learning. S-Net~\cite{slimnet} introduces a 4-width framework to incorporate different complexities into one network, and proposes the switchable batch normalization for slimmable training. \cite{multiexitkd} leverages knowledge distillation to train a multi-exit network. However, these approaches can only execute at a limited number of constraints. US-Net~\cite{uslimnet} can instantly adjust the runtime network width for arbitrary accuracy-efficiency trade-offs. 
%is the first to train a network with arbitrary widths. It proposes to compute post-training batch normalization statistics for arbitrary widths, and further introduces two improved training techniques to boost accuracy. 
But its performance degrades significantly as the budget lower bound goes down. \cite{onceforall} proposes progressive shrinking to finetune sub-networks from a well-trained large network, but the training process is complex and expensive.

% {\noindent\bf Knowledge Distillation.} Knowledge distillation \cite{kd} is another effective strategy for model compression. The idea is to transfer the knowledge from an over-parameterized teacher to a compressed student network. The student network learns from the ground truth label and the teacher's output. 
% %A series of works has been inspired from this idea. 
% FitNet \cite{romero2014fitnets} trains a thinner and deeper student network with hints, which are defined as the final output and intermediate representations from the teacher net. \cite{progressivefitnet} extends FitNet to iteratively training with hints. \cite{deepmutual} leverages multiple networks to learn knowledge mutually from each other. %\cite{fusionkd} proposes to learn from both sub-networks and the fused features.

{\noindent\bf Multi-scale Representation Learning.} The effectiveness of multi-scale representation has been explored in various tasks. FPN \cite{fpn} fuses pyramid features for object detection and segmentation. \cite{ke2017multigrid} proposes a multi-grid convolution to pass message across the scale space. HRNet \cite{hrnet} designs a multi-branch structure to exchange information across different resolutions. However, these works resort to the multi-branch fusion structure which is unfriendly to parallelization \cite{shufflenetv2}. Our method does not modify the network structure, and the learned multi-scale representation is not only from image scale but also from network scale.

\section{Methodology}
\subsection{Preliminary}

% Our framework leverages the training techniques in US-Net \cite{uslimnet}. Therefore, we first introduce these techniques in this section to make this paper self-contained. 

{\bf Sandwich Rule.} US-Net~\cite{uslimnet} trains a network that is executable at any resource constraint. The solution is to randomly sample several network widths for training and accumulate their gradients for optimization. However, the performance of the sub-networks is bounded by the smallest width (e.g., $0.25\times$) and the largest width (e.g., $1.0\times$). Thus, the \textit{sandwich rule} is introduced to sample the smallest and largest widths plus two random ones for each training iteration.

{\noindent\bf Inplace Distillation.} Knowledge distillation~\cite{kd} is an effective method to transfer knowledge from a teacher network to a student network. Following the \textit{sandwich rule}, since the largest network is sampled in each iteration, it is natural to use the largest network, which is supervised by the ground truth labels, as the teacher to guide smaller sub-networks in learning. This gives a better performance than training all sub-networks with ground truth labels.

{\noindent\bf Post-statistics of Batch Normalization (BN).} US-Net proposes that each sub-network needs their own BN statistics (mean and variance), but it is insufficient to store the statistics of all the sub-networks. Therefore, US-Net collects BN statistics for the desired sub-network after training. Experimental results show that 2,000 samples are sufficient to get accurate BN statistics.

\subsection{Rethinking Efficient Network Design}
\label{sec2}
The computation cost of a vanilla convolution is $C_1 \times C_2 \times K \times K \times H \times W$, where $C_1$ and $C_2$ are the number of input and output channels, $K$ is the kernel size, $H$ and $W$ are output feature map sizes. Most previous works only focus on reducing $C_1 \times C_2$. The most widely used group convolution decomposes standard convolution into groups to reduce the computation to $C_1 \times (C_2/g) \times K \times K \times H \times W$, where $g$ is the number of groups. A larger $g$ gives a lower computation but leads to higher memory access cost (MAC) \cite{shufflenetv2}, making the network inefficient in practical applications. Pruning methods \cite{unstrprune, strprune, structuredprune2} also only consider reducing structure redundancies.
% while ignoring the redundancies in the input. 

In our approach, we shift the attention to reducing $H \times W$, i.e., lowering input resolution for the following reasons. First, as demonstrated in Fig. \ref{figure1}, balancing between width and resolution achieves better accuracy-efficiency trade-offs. Second, downsampling input resolution does not necessarily hurt the performance. It even sometimes benefits the performance. \cite{chin2019adascale} points out that lower image resolution may produce better detection accuracy by reducing focus on redundant details. Third, different resolutions contain different information \cite{octave}. Lower resolution images may contain more global structures while higher resolution ones may encapsulate more fine-grained patterns. Learning multi-scale representations from different scaled images and features has been proven effective in previous works \cite{fpn, ke2017multigrid, hrnet}. But these methods resort to a multi-branch structure which is unfriendly to parallelization \cite{shufflenetv2}. Motivated by these observations, we propose a mutual learning framework to consider network scale and input resolution simultaneously for effective network accuracy-efficiency trade-offs.

\subsection{Mutual Learning Framework}

{\bf Sandwich Rule and Mutual Learning.}
As discussed in Section \ref{sec2}, different resolutions contain different information. We want to take advantage of this attribute to learn robust representations and better width-resolution trade-offs. 
%In US-Net, \textit{sandwich rule} is proposed to train a network that is executable at any width. But it can also be viewed as a scheme of mutual learning \cite{deepmutual} where an ensemble of networks are learning collaboratively. 
The \textit{sandwich rule} in US-Net can be viewed as a scheme of mutual learning \cite{deepmutual} where an ensemble of networks are learned collaboratively. 
Since the sub-networks share weights with each other and are optimized jointly, they can transfer their knowledge to each other. Larger networks can take advantage of the features captured by smaller networks. Also, smaller networks can benefit from the stronger representation ability of larger networks. In light of this, we feed each sub-network with different input resolutions. By sharing knowledge, each sub-network is able to capture multi-scale representations.

{\noindent\bf Model Training.} We present an example to illustrate our framework in Fig. \ref{slimnet}. We train a network where its width ranges from $0.25\times$ to $1.0\times$. We first follow the \textit{sandwich rule} to sample four sub-networks, i.e., the smallest ($0.25\times$), the largest ($1.0\times$) and \textit{two random width ratios} $\alpha_1, \alpha_2 \in (0.25, 1)$. Then, unlike traditional ImageNet training with $224\times 224$ input, we resize the input image to four resolutions \{224, 196, 160, 128\} and feed them into different sub-networks. We denote the weights of a sub-network as $W_{0:w}$, where $w \in (0, 1]$ is the width of the sub-network and $0:w$ means the sub-network adopts the first $w \times 100\%$ weights of each layer of the full network. $I_{R=r}$ represents a $r\times r$ input image. Then $N(W_{0:w},I_{R=r})$ represents the output of a sub-network with width $w$ and input resolution $r\times r$. For the largest sub-network (i.e., the full-network in Fig.~\ref{slimnet}), we always train it with the highest resolution ($224\times 224$) and ground truth label $y$. The loss for the full network is 
\begin{equation}
    loss_{full} = CrossEntropy(N(W_{0:1},I_{R=224}),\, y).
\end{equation}
For the other sub-networks, we randomly pick an input resolution from \{224, 196, 160, 128\} and train it with the output of the full-network. The loss for the $i$-th sub-network is
\begin{equation}
    loss_{sub_i} = KLDiv(N(W_{0:w_i},I_{R=r_i}),\, N(W_{0:1},I_{R=224})),
\end{equation}
where $KLDiv$ is the Kullback-Leibler divergence. The total loss is the summation of the full-network and sub-networks, i.e.,
\begin{equation}
    loss = loss_{full}+\sum_{i=1}^3loss_{sub_i}.
\end{equation}
The reason for training the full-network with the highest resolution is that the highest resolution contains more details. Also, the full-network has the strongest learning ability to capture the  discriminatory information from the image data.

\begin{figure}[t]
\centering
    \includegraphics[width=1.0\linewidth]{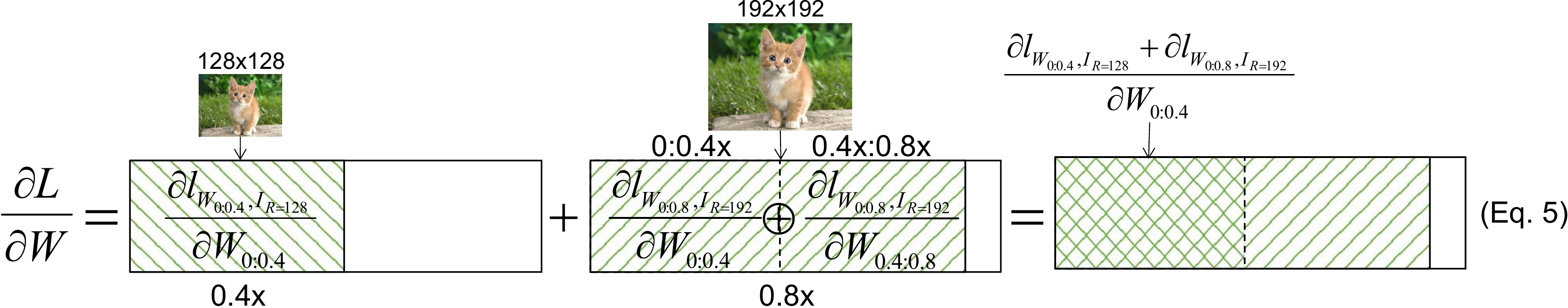}
    \caption{\small Illustration of the mutual learning from network width and input resolution.}
    \label{mutual}
\end{figure}

{\noindent\bf Mutual Learning from Width and Resolution.} In this part, we explain why the proposed framework can mutually learn from different widths and resolutions. For ease of demonstration, we only consider two network widths $0.4\times$ and $0.8\times$, and two resolutions 128 and 192 in this example. As shown in Fig. \ref{mutual}, sub-network $0.4\times$ selects input resolution 128, sub-network $0.8\times$ selects input resolution 192. Then we can define the gradients for sub-network $0.4\times$ and $0.8\times$ as $\frac{\partial l_{W_{0:0.4},I_{R=128}}}{\partial W_{0:0.4}}$ and $\frac{\partial l_{W_{0:0.8},I_{R=192}}}{\partial W_{0:0.8}}$, respectively. Since sub-network $0.8\times$ shares weights with $0.4\times$, we can decompose its gradient as 
\begin{equation}
\small
\begin{aligned}
    \frac{\partial l_{W_{0:0.8},I_{R=192}}}{\partial W_{0:0.8}}=\frac{\partial l_{W_{0:0.8},I_{R=192}}}{\partial W_{0:0.4}} \oplus \frac{\partial l_{W_{0:0.8},I_{R=192}}}{\partial W_{0.4:0.8}}
\end{aligned}
\end{equation}
where $\oplus$ is vector concatenation. Since the gradients of the two sub-networks are accumulated during training, the total gradients are computed as
\begin{equation}
\small
\begin{aligned}
    \frac{\partial L}{\partial W} &=\frac{\partial l_{W_{0:0.4},I_{R=128}}}{\partial W_{0:0.4}} + \frac{\partial l_{W_{0:0.8},I_{R=192}}}{\partial W_{0:0.8}} \\
    &= \frac{\partial l_{W_{0:0.4},I_{R=128}}}{\partial W_{0:0.4}} + \left(\frac{\partial l_{W_{0:0.8},I_{R=192}}}{\partial W_{0:0.4}} \oplus \frac{\partial l_{W_{0:0.8},I_{R=192}}}{\partial W_{0.4:0.8}}\right) \\
    &= \frac{\partial l_{W_{0:0.4},I_{R=128}} + \partial l_{W_{0:0.8},I_{R=192}}}{\partial W_{0:0.4}} \oplus \frac{\partial l_{W_{0:0.8},I_{R=192}}}{\partial W_{0.4:0.8}}
\end{aligned}
\end{equation}

% \begin{equation}
% \begin{aligned}
%     grad &= grad_{W_{0:0.3,},I_{R=128}} + grad_{W_{0:0.8,},I_{R=192}} \\
%     &= (grad_{W_{0:0.3,},I_{R=128}} + grad_{W_{0:0.3,},I_{R=192}}) + grad_{W_{0.3:0.8,},I_{R=192}}.
% \end{aligned}
% \end{equation}
\noindent Therefore, the gradient for sub-network $0.4\times$ is $\frac{\partial l_{W_{0:0.4},I_{R=128}} + \partial l_{W_{0:0.8},I_{R=192}}}{\partial W_{0:0.4}}$, and it consists of two parts. The first part is derived from itself (0 : 0.4$\times$) with 128 input resolution. The second part is derived from sub-network $0.8\times$ (i.e., $0:0.4\times$ portion) with 192 input resolution. Thus the sub-network is able to capture multi-scale representations from different input resolutions and network scales. Due to the random sampling of network width, every sub-network is able to learn multi-scale representations in our framework.

%{\bf Deploy on hardware.} 
{\noindent\bf Model Inference.} The trained model is executable at various width-resolution configurations. The goal is to find the best configuration under a particular resource constraint. A simple way to achieve this is via a query table. Specifically, we sample network width from $0.25\times$ to $1.0\times$ with a step-size of $0.05\times$, and sample network resolution from \{224, 192, 160, 128\}. We test all these width-resolution configurations on a validation set and choose the best one under a given constraint (FLOPs or latency). Since there is no re-training, the whole process is once for all. 
%It can be done before deployment and is once for all.
\section{Experiments}

In this section, we first present our results on ImageNet \cite{imagenet} classification to illustrate the effectiveness of MutualNet. Next, we conduct extensive ablation studies to analyze the mutual learning scheme. Finally, we apply MutualNet to transfer learning datasets and COCO \cite{coco} object detection and instance segmentation to demonstrate its robustness and generalization ability. 
% We conduct extensive experiments to illustrate the effectiveness of the proposed framework. Next, we fine-tune the pre-trained model on popular transfer learning datasets to demonstrate the robustness and generalization ability of the learned representations. We further apply our framework to COCO object detection and instance segmentation. \textbf{To the best of our knowledge, we are the first to benchmark arbitrary-constraint dynamic networks on detection and instance segmentation.}
\subsection{Evaluation on ImageNet Classification}
\label{evalImageNet}
We compare our MutualNet with US-Net and independently-trained networks on the ImageNet dataset. 
We evaluate our framework on two popular light-weight structures, MobileNet v1 \cite{howard2017mobilenets} and MobileNet v2 \cite{mobilenetv2}. 
These two networks also represent non-residual and residual structures respectively.

{\noindent\bf Implementation Details.} We compare with US-Net under the \textbf{same} dynamic FLOPs constraints ([13, 569] MFLOPs on MobileNet v1 and [57, 300] MFLOPs on MobileNet v2). US-Net uses width scale [0.05, 1.0]$\times$ on MobileNet v1 and [0.35, 1.0]$\times$ on MobileNet v2 based on the 224$\times$224 input resolution. \textbf{To meet the same dynamic constraints}, our method uses width scale [0.25, 1.0]$\times$ on MobileNet v1 and [0.7, 1.0]$\times$ on MobileNet v2 with downsampled input resolutions \{224, 192, 160, 128\}. Due to the lower input resolutions, our method is able to use higher width lower bounds (i.e., $0.25\times$ and $0.7\times$) than US-Net. The other training settings are the same as US-Net.
    
\begin{figure}[t]
    % \vspace{-\intextsep}
    \centering
    \subfigure[MobileNet v1 backbone]{
    \includegraphics[width=0.45\textwidth]{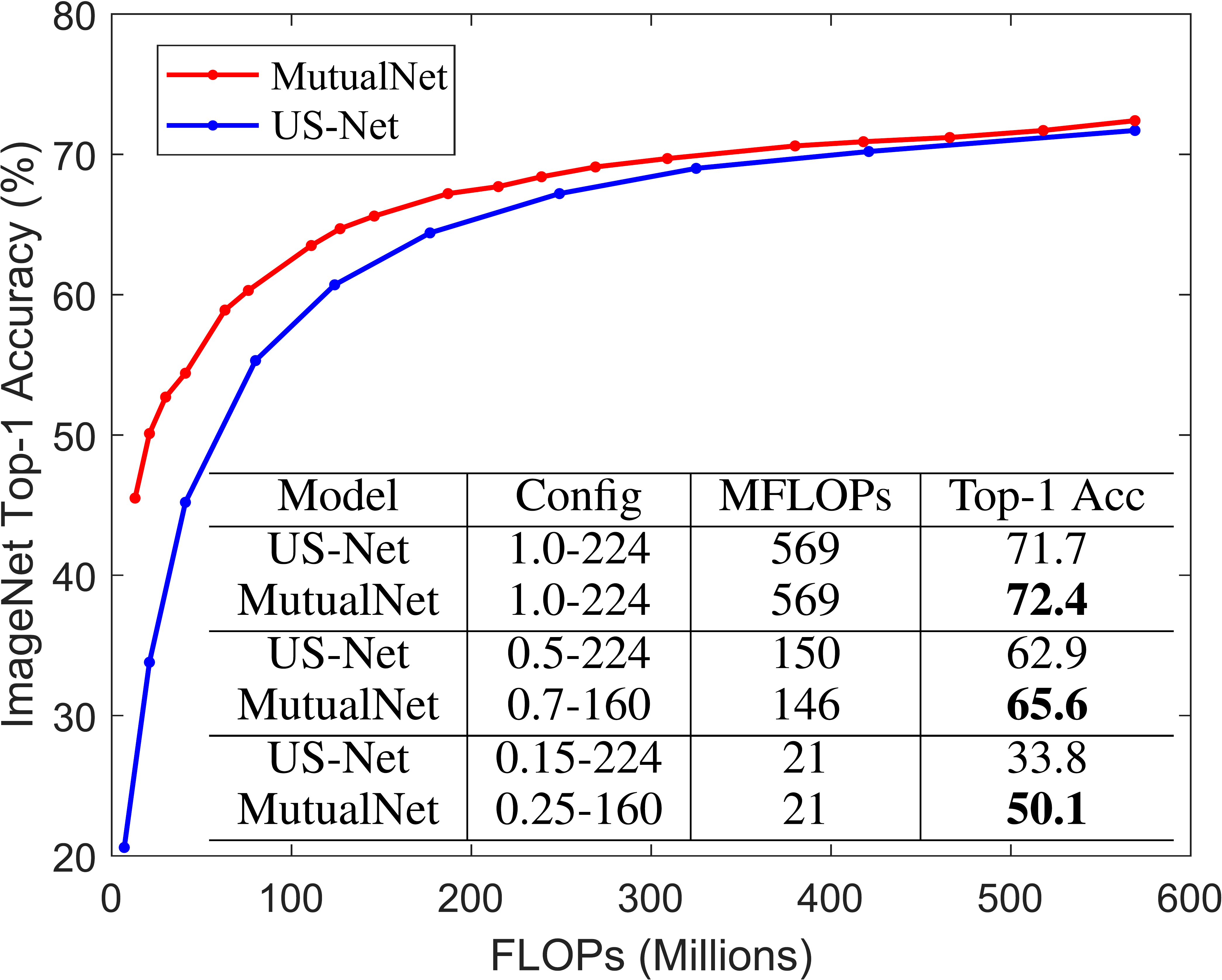} 
}
    \subfigure[MobileNet v2 backbone]{
    \includegraphics[width=0.45\textwidth]{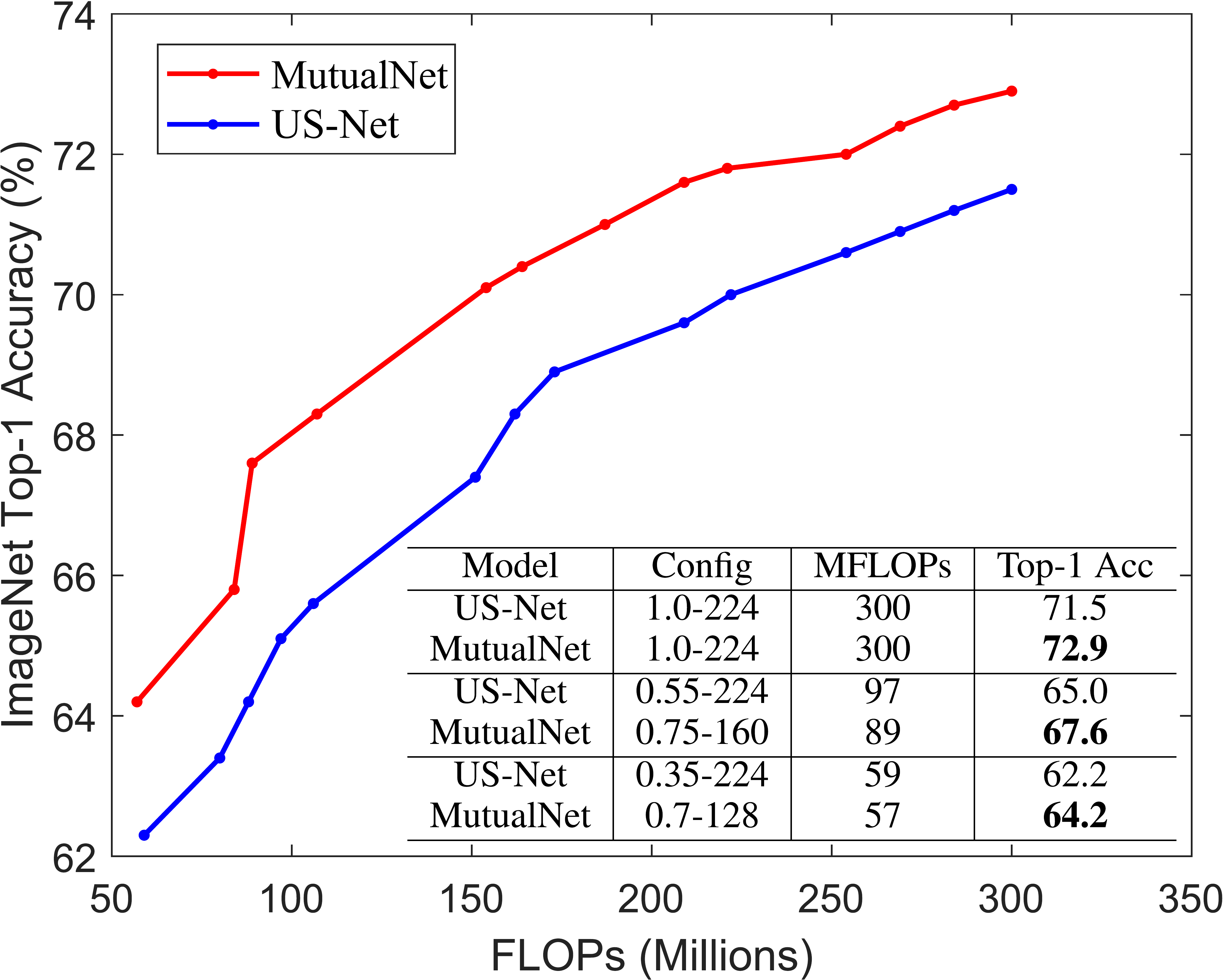}
} 
    \vspace{-0.5em}
    \caption{Accuracy-FLOPs curves of our proposed MutualNet and US-Net. (a) is based on MobileNet v1 backbone. (b) is based on MobileNet v2 backbone.}
    \label{USNet-ours}
\end{figure}

{\noindent\bf Comparison with US-Net.} We first compare our framework with US-Net on MobileNet v1 and MobileNet v2 backbones. The Accuracy-FLOPs curves are shown in Fig. \ref{USNet-ours}. We can see that our framework consistently outperforms US-Net on both MobileNet v1 and MobileNet v2 backbones. Specifically, we achieve significant improvements under small computation costs. This is because our framework considers both network width and input resolution, and can find a better balance between them. For example, if the resource constraint is 150 MFLOPs, US-Net has to reduce the width to $0.5\times$ given its constant input resolution $224$, while our MutualNet can meet this budget by a balanced configuration of ($0.7\times$ - 160), leading to a better accuracy (65.6\% (Ours) vs. 62.9\% (US-Net) as listed in the table of Fig. \ref{USNet-ours}(a)). On the other hand, our framework is able to learn multi-scale representations which further boost the performance of each sub-network. We can see that even for the same configuration (e.g., 1.0$\times$-224) our approach clearly outperforms US-Net, i.e., 72.4\% (Ours) vs. 71.7\% (US-Net) on MobileNet v1, and 72.9\% (Ours) vs. 71.5\% (US-Net) on MobileNet v2 (Fig. \ref{USNet-ours}).

\begin{figure}[h]
    \centering
    \subfigure[MobileNet v1 backbone]{
    \includegraphics[width=0.45\textwidth]{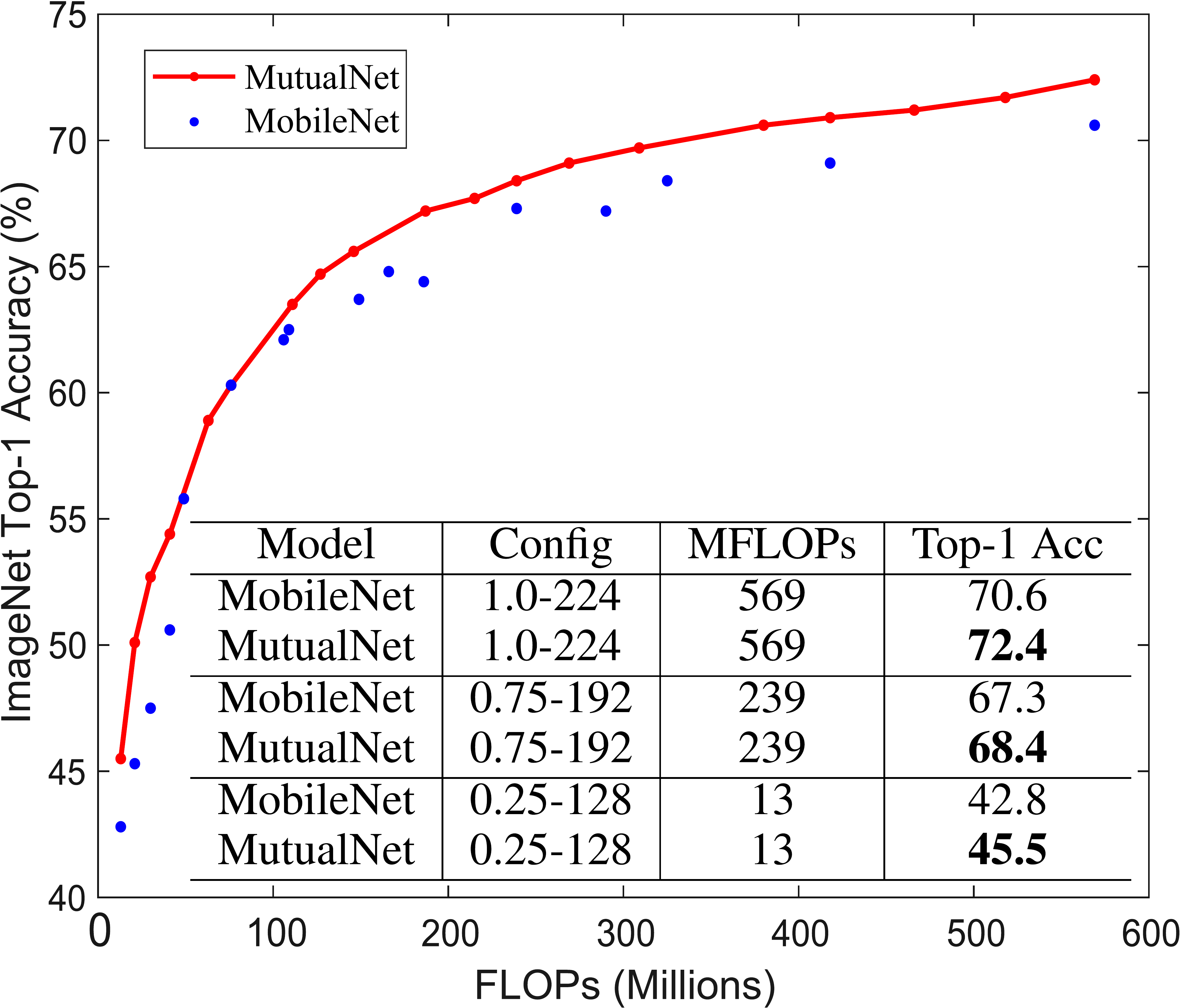} 
}
    \subfigure[MobileNet v2 backbone]{
    \includegraphics[width=0.45\textwidth]{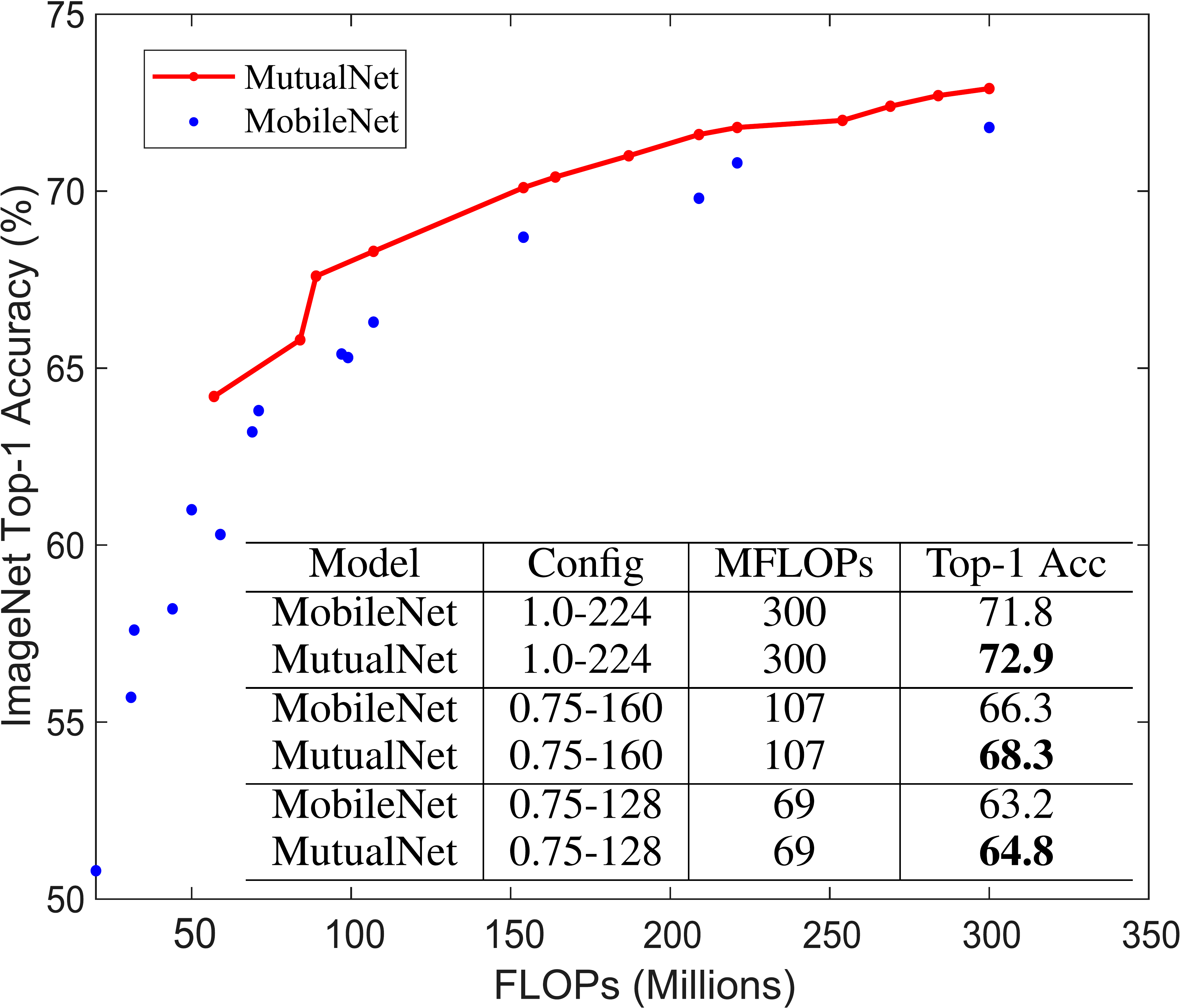}
}
    \caption{Accuracy-FLOPs curves of our MutualNet and independently-trained MobileNets. (a) is MobileNet v1 backbone. (b) is MobileNet v2 backbone. The results for different MobileNets configurations are taken from the papers \cite{howard2017mobilenets, mobilenetv2}.}
    % \vspace{-\intextsep}
\label{MNet-reso}
\end{figure}

{\noindent\bf Comparison with Independently Trained Networks.} Different scaled MobileNets are trained separately in \cite{howard2017mobilenets, mobilenetv2}. The authors consider width and resolution as independent factors, thus cannot leverage the information contained in different configurations. We compare the performance of MutualNet with independently-trained MobileNets under different width-resolution configurations in Fig. \ref{MNet-reso}. For MobileNet v1, widths are selected from \{$1.0\times$, $0.75\times$, $0.5\times$, $0.25\times$\}, and resolutions are selected from \{224, 192, 160, 128\}, leading to 16 configurations in total. Similarly, MobileNet v2 selects configurations from \{1.0$\times$, 0.75$\times$, 0.5$\times$, 0.35$\times$\} and \{224, 192, 160, 128\}. From Fig. \ref{MNet-reso}, our framework consistently outperforms MobileNets. Even for the same width-resolution configuration (although it may not be the best configuration MutualNet finds at that specific constraint), MutualNet can achieve much better performance. This demonstrates that MutualNet not only finds the better width-resolution balance but also learns stronger representations by the mutual learning scheme.

\begin{figure}[t]
    \centering
    % \vspace{-\intextsep}
    \includegraphics[width=0.95\textwidth]{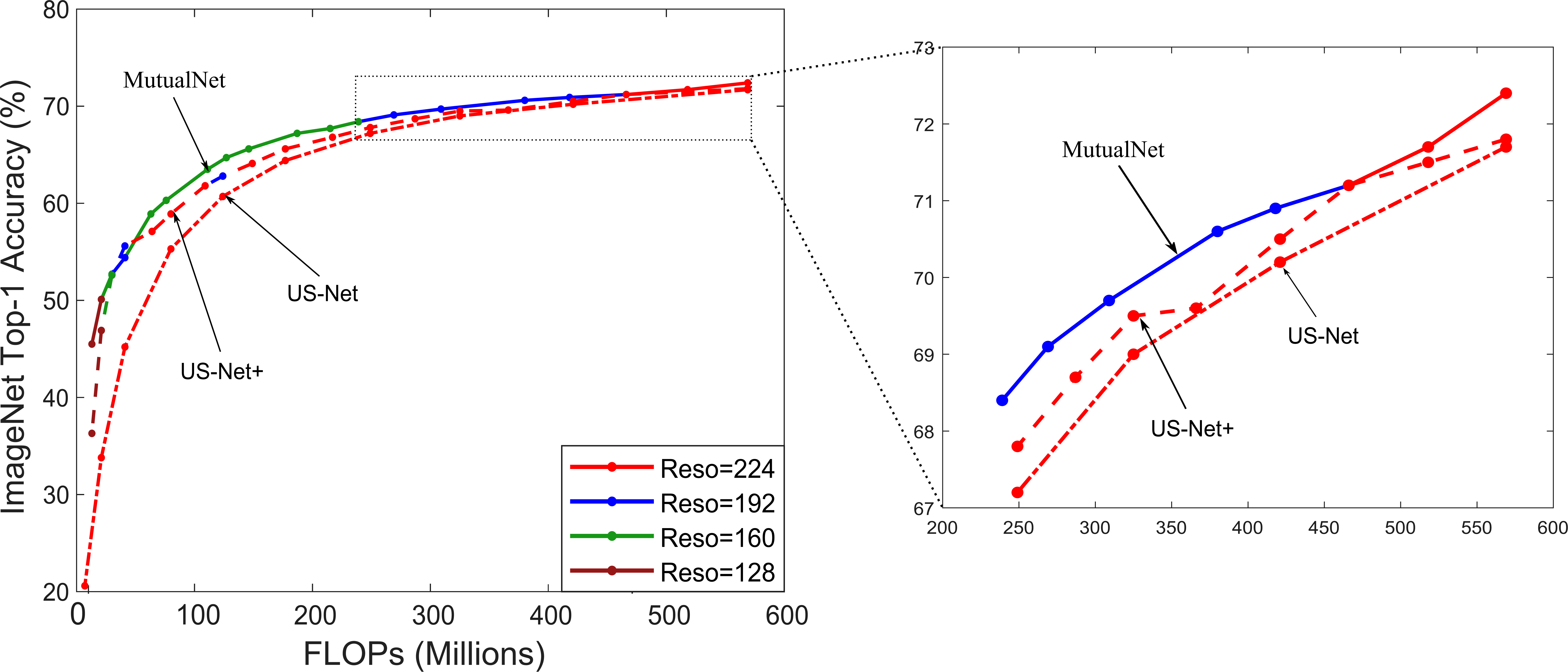} 
    \caption{The Accuracy-FLOPs curves are based on MobileNet v1 backbone. We highlight the selected resolution under different FLOPs with different colors. For example, the solid green line indicates when the constraint range is [41, 215] MFLOPs, our method constantly selects input resolution 160 but reduces the width to meet the resource constraint. Best viewed in color.}
    \label{Ours-US-Net+}
\end{figure}

\subsection{Ablation Study}
\subsubsection{Balanced Width-Resolution Configuration via Mutual Learning.} 

As evident in Fig. \ref{figure1}, we can apply different resolutions to US-Net \textit{during inference} to yield improvement over the original US-Net. However, this cannot achieve the optimal width-resolution balance due to lack of width-resolution mutual learning. In the experiment, we test US-Net at width scale [0.25, 1.0]$\times$ with input resolutions \{224, 192, 160, 128\} and denote this improved model as US-Net+. In Fig. \ref{Ours-US-Net+}, we plot the Accuracy-FLOPs curves of our method and US-Net+ based on MobileNet v1 backbone, and highlight the selected input resolutions with different colors. As we decrease the FLOPs ($569\rightarrow468$ MFLOPs), our MutualNet first reduces network width to meet the constraint while keeping the 224$\times$224 resolution (red line in Fig. \ref{Ours-US-Net+}). After 468 MFLOPs, MutualNet selects lower input resolution (192) and continues reducing the width to meet the constraint. On the other hand, US-Net+ cannot find such balance. It always slims the network width and uses the same (224) resolution as the FLOPs decreasing until it goes to really low. This is because US-Net+ does not incorporate input resolution into the learning framework. Simply applying different resolutions during inference cannot achieve the optimal width-resolution balance.

\subsubsection{Difference with EfficientNet.}

EfficientNet acknowledges the importance of balancing between network width, depth and resolution. But they are considered as independent factors. The authors use grid search over these three dimensions and train each configuration independently to find the optimal one under certain constraint, while our MutualNet incorporates width and resolution in a unified framework. We compare with the best model scaling EfficientNet finds for MobileNet v1 at 2.3 BFLOPs (scale up baseline by 4.0$\times$).  Similar to this scale setting, we train our framework with a width range of [$1.0\times,2.0\times$] (scaled by [1.0$\times$,4.0$\times$]), and select resolutions from \{224, 256, 288, 320\}. This makes MutualNet executable in the range of [0.57, 4.5] BFLOPs. We pick the best performed width-resolution configuration at 2.3 BFLOPs. The results are compared in Table \ref{table1}. MutualNet achieves significantly better performance than EfficientNet because it can capture multi-scale representations for each model scaling due to the width-resolution mutual learning.

\begin{table}[t]
\caption{ImageNet Top-1 accuracy on MobileNet v1 backbone. $d$: depth, $w$: width, $r$: resolution.}
\begin{center}
\begin{tabular}{c|c|c|c}
    \hline
     Model& Best Model Scaling& FLOPs& Top-1 Acc  \\
     \hline
     EfficientNet \cite{efficientnet}& $d=1.4, w=1.2, r=1.3$& 2.3B& 75.6\% \\
     MutualNet& $w=1.6, r=1.3$& 2.3B& {\bf 77.1\%} \\
     \hline
\end{tabular}
\end{center}
\vspace{-\intextsep}
\label{table1}
\end{table}

% \begin{figure}[h]
%     \centering
%     \subfigure[Cifar 100]{
%     \includegraphics[width=0.31\textwidth]{fig/cifar100c.pdf} 
% }
%     \subfigure[Food 101]{
%     \includegraphics[width=0.31\textwidth]{fig/food101c.pdf}
% }
%     \subfigure[MIT Indoor 67]{
%     \includegraphics[width=0.31\textwidth]{fig/mit67c.pdf}
% }
%     \caption{Accuracy-FLOPs curves on different transfer learning datasets.}
%     \label{transfer}
% \end{figure}

\subsubsection{Difference with Multi-scale Data Augmentation.}
Multi-scale data augmentation is popular in detection and segmentation. We show that, first, our method is different from multi-scale data augmentation in principle. Second, our method significantly outperforms multi-scale data augmentation.

In multi-scale data augmentation, 
%we feed networks with different scale images in each iteration. 
the network may take images of different resolutions in different iterations. 
But within each iteration, the network weights are optimized in the same resolution direction. While our method randomly samples four sub-networks which share weights with each other. Since sub-networks can select different image resolutions, the weights are optimized in a mixed resolution direction in each iteration as illustrated in Fig. \ref{mutual}. This enables each sub-network to effectively learn multi-scale representations from width and resolution. To validate the superiority of our mutual learning scheme, we apply multi-scale data augmentation to MobileNet and US-Net.

\textit{MobileNet + Multi-scale data augmentation.} We train MobileNet v2 ($1.0\times$ width) with multi-scale images. To have a fair comparison, input images are randomly sampled from scales \{224, 192, 160, 128\} and the other settings are the same as MutualNet. As shown in Table \ref{mnetmultiscale}, multi-scale data augmentation only marginally improves the baseline (MobileNet v2) while our MutualNet (MobileNet v2 backbone) clearly outperforms both of them by considerable margins.

\iffalse
\begin{table}[t]
\caption{Comparison between MutualNet and multi-scale training.}
\begin{center}
\begin{tabular}{c|c|c|c}
    \hline
     MobileNet v2 (1.0$\times$ - 224)& Baseline & Multi-scale training& MutualNet \\
     \hline
     ImageNet Top-1 Acc& 71.8& 72.0& {\bf72.9} \\
     \hline
\end{tabular}
\end{center}
% \vspace{-\intextsep}
\label{mnetmultiscale}
\end{table}
\fi

\begin{table}[t]
\caption{Comparison between MutualNet and multi-scale data augmentation.}
\begin{center}
\begin{tabular}{c|c}
    \hline
     Model& ImageNet Top-1 Acc  \\
     \hline
     MobileNet v2 (1.0$\times$ - 224) - Baseline & 71.8\% \\
     Baseline + Multi-scale data augmentation & 72.0\% \\
     MutualNet (MobileNet v2 backbone) & {\bf72.9\%}\\
     \hline
\end{tabular}
\end{center}
% \vspace{-\intextsep}
\label{mnetmultiscale}
\end{table}

\begin{figure}[t]
\centering
\begin{minipage}[]{0.45\linewidth}
\centering
    % \vspace{-\intextsep}
    \includegraphics[width=0.95\linewidth]{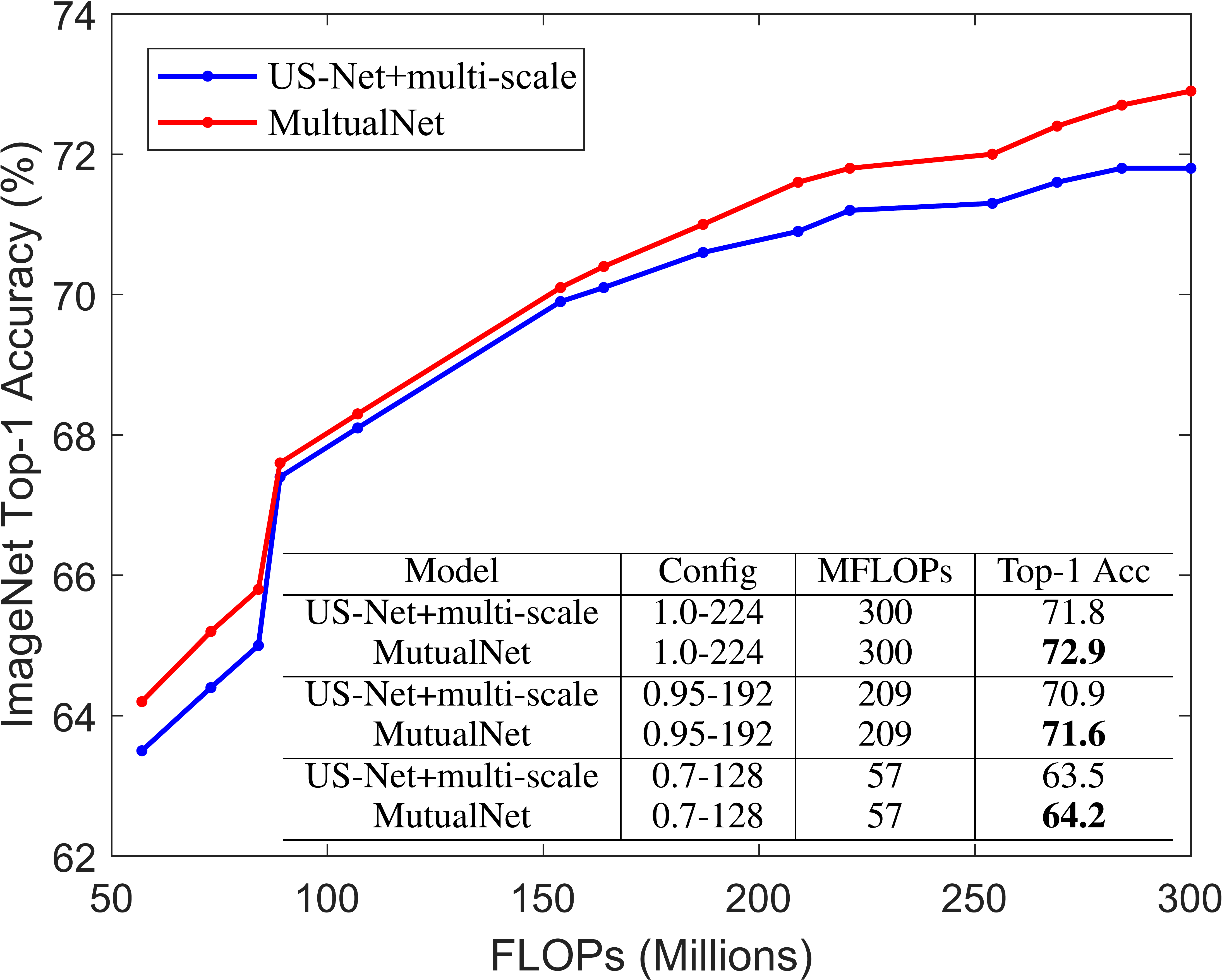} 
    \caption{MutualNet and US-Net + multi-scale data augmentation.}
    \label{USmultiscale}
\end{minipage}
\hspace{0.1in}
\begin{minipage}[]{0.45\linewidth}
\centering
\includegraphics[width=0.95\linewidth]{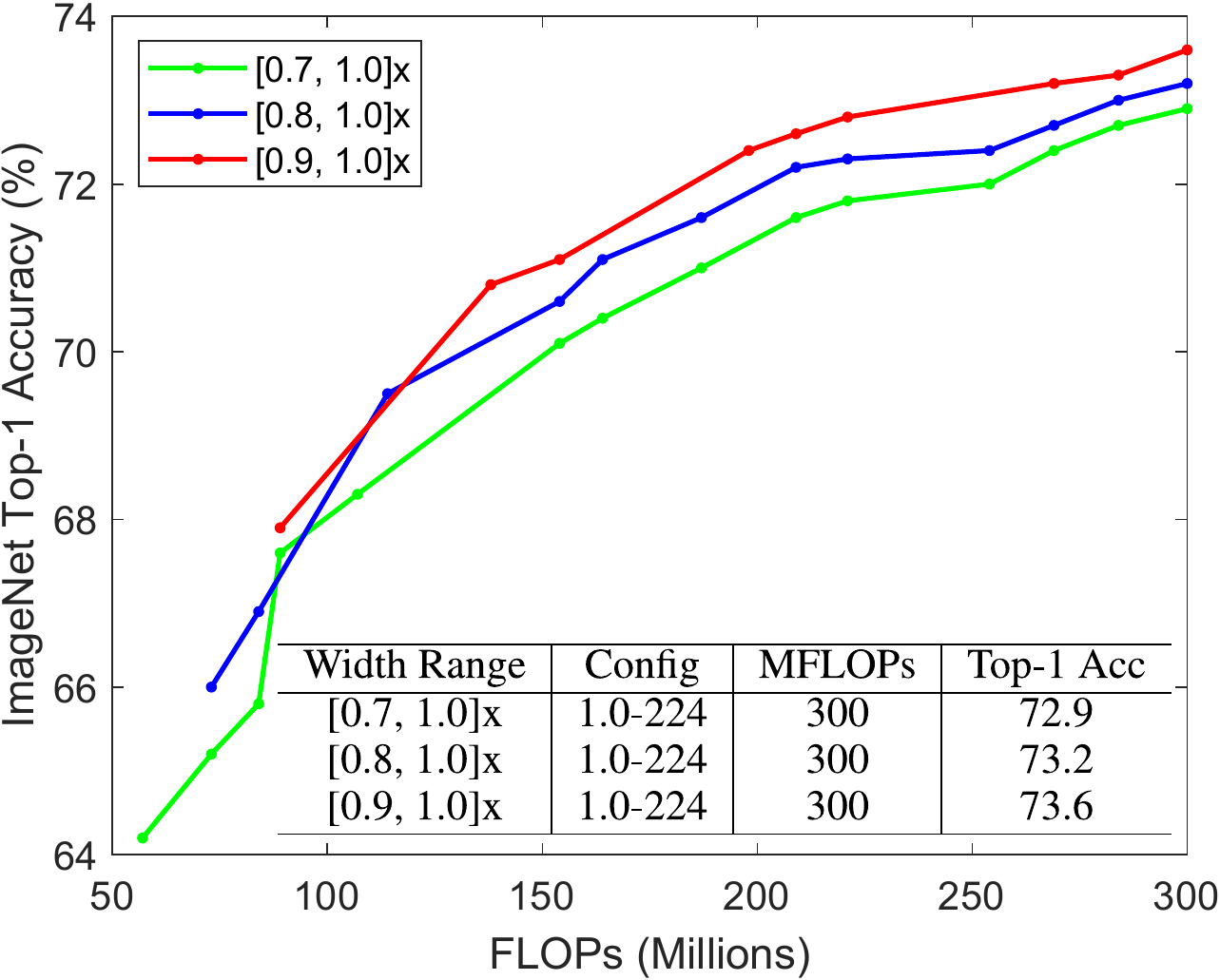} 
    \caption{Accuracy-FLOPs curves of different width lower bounds.}
    \label{lowerbound}
\end{minipage}
% \vspace{-\intextsep}
\end{figure}

\textit{US-Net + Multi-scale data augmentation.} Different from our framework which feeds different scaled images to different sub-networks, in this experiment, we \textit{randomly} choose a scale from \{224, 192, 160, 128\} and feed the \textit{same} scaled image to all sub-networks in each iteration. That is each sub-network takes the same image resolution. In this way, the weights are still optimized towards a single resolution direction in each iteration, but the direction varies among different iterations. The experiment is based on MobileNet v2. Width and resolution settings are the same as those in Sec. \ref{evalImageNet}. As shown in Fig. \ref{USmultiscale}, our method clearly performs better than \textit{US-Net + multi-scale data augmentation} over the entire FLOPs spectrum. \textit{These experiments demonstrate that the improvement comes from our mutual learning scheme rather than the multi-scale data augmentation.}

\subsubsection{Effects of Width Lower Bound.} The executable constraint range and model performance are affected by the width lower bound. To study its effects, we conduct experiments with three different lower bounds (0.7$\times$, 0.8$\times$, 0.9$\times$) on MobileNet v2. The results in Fig. \ref{lowerbound} show that a higher lower bound gives better overall performance, but the executable range is narrower. 
%This is consistent with the results in US-Net \cite{uslimnet}. 
One interesting observation is that the performance of the full-network (1.0$\times$-224) is also largely improved as the width lower bound increases from 0.7$\times$ to 0.9$\times$. This property is not observed in US-Net. We attribute this to the robust and well-generalized multi-scale representations which can be effectively re-used by the full-network, while in US-Net, the full-network cannot effectively benefit from %improved
sub-networks.

% \begin{figure}[h]
%     \centering
%     % \vspace{-\intextsep}
%     \includegraphics[width=0.7\textwidth]{fig/showbalance.pdf} 
%     \caption{Placeholder}
%     \label{lowerbound}
% \end{figure}
\begin{table}[t]
\centering
\begin{minipage}[]{0.44\linewidth}
\begin{center}
\caption{Top-1 Accuracy (\%) on Cifar-10 and Cifar-100.}
\begin{tabular}{|c|c|c|c|}
    \hline
     \makecell[c]{WideResNet \\ -28-10}& \makecell[c]{GPU search \\ hours}& C-10& C-100\\
     \hline
     Baseline& \textbf{0}& 96.1& 81.2 \\
     \hline 
     Cutout \cite{cutout}& \textbf{0}& 96.9& 81.6 \\
     \hline
     AA \cite{cubuk2019autoaugment}& 5000& \textbf{97.4}& 82.9 \\
     \hline
     Fast AA \cite{fastautoaug}& 3.5& 97.3& 82.7 \\
     \hline
     MutualNet & \textbf{0}& 97.2& \textbf{83.8} \\
     \hline
\end{tabular}
\label{boostsingle-cifar}
\end{center}
\end{minipage}
% \hspace{0.1in}
\begin{minipage}[]{0.55\linewidth}
  \centering
  \caption{Top-1 Accuracy (\%) on ImageNet.}
  \begin{tabular}{|c|c|c|}
    \hline
     ResNet-50& Additional Cost& Top-1 Acc  \\
     \hline
     Baseline& \textbackslash& 76.5 \\
     \hline 
     Cutout \cite{cutout}& \textbackslash& 77.1 \\
     \hline
     KD \cite{kdimprovement}& Teacher Network& 76.5 \\
     \hline
     SENet \cite{senet}& SE block& 77.6 \\
     \hline
     AA \cite{cubuk2019autoaugment}& 15000 GPU hours& 77.6 \\
     \hline
     Fast AA \cite{fastautoaug}& 450 GPU hours& 77.6 \\
     \hline
     MutualNet & \textbackslash& \textbf{78.6} \\
     \hline
  \end{tabular}
  \label{boostsingle-imagenet}
\end{minipage}
% \vspace{-0.2cm}
\end{table}

\subsubsection{Boosting Single Network Performance.} As discussed above, the performance of the full-network is greatly improved as we increase the width lower bound. Therefore, we apply our training framework to improve the performance of a single full network. We compare our method with the popular performance-boosting techniques (e.g., AutoAugmentation (AA) \cite{fastautoaug,cubuk2019autoaugment}, SENet \cite{senet} and Knowledge Distillation (KD) \cite{kdimprovement}) to show its superiority. We conduct experiments using Wide-ResNet \cite{wideresnet} on Cifar-10 and Cifar-100 and ResNet-50 on ImageNet. MutualNet adopts the width range [0.9, 1.0]$\times$ as it achieves the best-performed full-network in Fig. \ref{lowerbound}. The resolution is sampled from \{32, 28, 24, 20\} on Cifar-10 and Cifar-100 and \{224, 192, 160, 128\} on ImageNet. Wide-ResNet is trained for 200 epochs following \cite{wideresnet}. ResNet is trained for 100 epochs. The results are compared in Table \ref{boostsingle-cifar} and Table \ref{boostsingle-imagenet}. MutualNet achieves substantial improvements over other techniques. It is important to note that MutualNet is a \textbf{general training scheme} which does not need the expensive searching procedure or additional network blocks or stronger teacher networks. Moreover, MutualNet training is as easy as the regular training process, and is orthogonal to other performance-boosting techniques, e.g., AutoAugmentation \cite{fastautoaug,cubuk2019autoaugment}. Therefore, it can be easily combined with those methods.% for enhanced performance.

% Our framework clearly outperforms Cutout on Cifar-10, Cifar-100 and ImageNet. Compared to AutoAug \cite{cubuk2019autoaugment} and Fast AutoAug \cite{fastautoaug}, our framework achieves comparable performance on Cifar-10 and significant improvement on Cifar-100 and ImageNet. Moreover, our framework is as easy as the standard training procedure, and it does not include the expensive searching process (GPU hours: 15000 (AA) vs. \textbf{0} (MutualNet)). Therefore, the proposed framework is promising to serve as a \textbf{plug-and-play} training strategy to boost the performance of single network.

\subsection{Transfer Learning}
\begin{figure}[t]
    \centering
    % \vspace{-\intextsep}
    \includegraphics[width=0.98\linewidth]{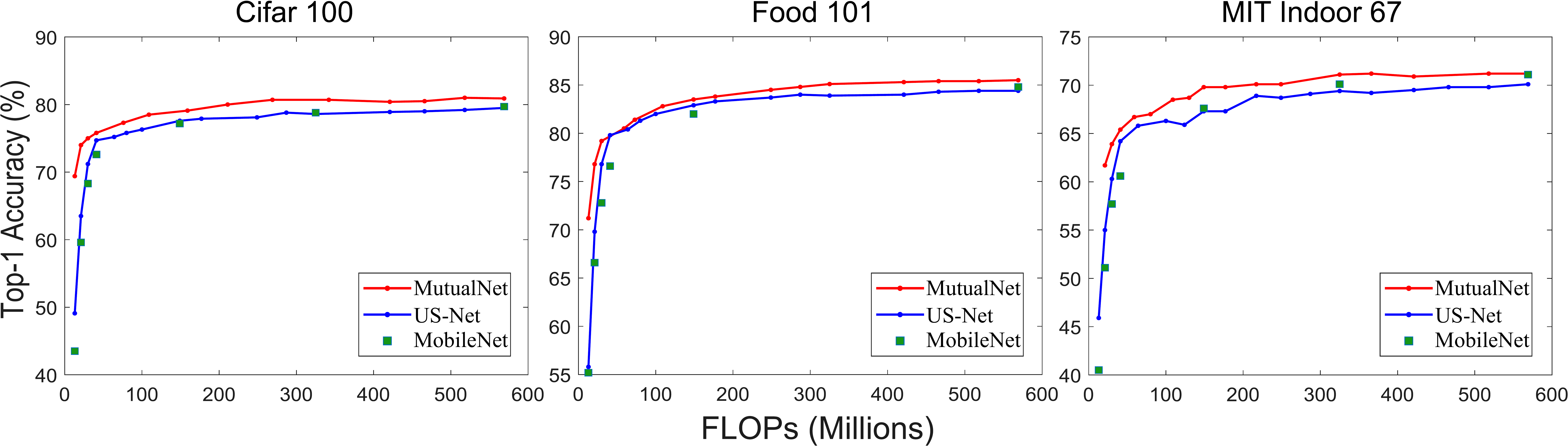} 
    \caption{Accuracy-FLOPs curves on different transfer learning datasets.}
    \label{transfer}
\end{figure}

To evaluate the representations learned by our method, we further conduct experiments on three popular transfer learning datasets, Cifar-100 \cite{cifar}, Food-101 \cite{food101} and MIT-Indoor67 \cite{MITindoor67}. Cifar-100 is superordinate-level object classification, Food-101 is fine-grained classification and MIT-Indoor67 is scene classification. Such large variety is suitable to evaluate the robustness of the learned representations. We compare our approach with US-Net and MobileNet v1. We fine-tune ImageNet pre-trained models
\footnote{For US-Net, we use the officially released model at width [0.25, 1.0]$\times$.} 
with a batch size of 256, initial learning rate of 0.1 with cosine decay schedule and a total of 100 epochs. Both MutualNet and US-Net are trained with width range [0.25, 1.0]$\times$ and tested with resolutions from \{224, 192, 160, 128\}.
The results are shown in Fig. \ref{transfer}. Again, our MutualNet achieves consistently better performance than US-Net and MobileNet. This verifies that MutualNet is able to learn well-generalized representations.

\subsection{Object Detection and Instance Segmentation}
\begin{figure}[t]
    \centering
    \subfigure[Object Detection]{
    \includegraphics[width=0.425\textwidth]{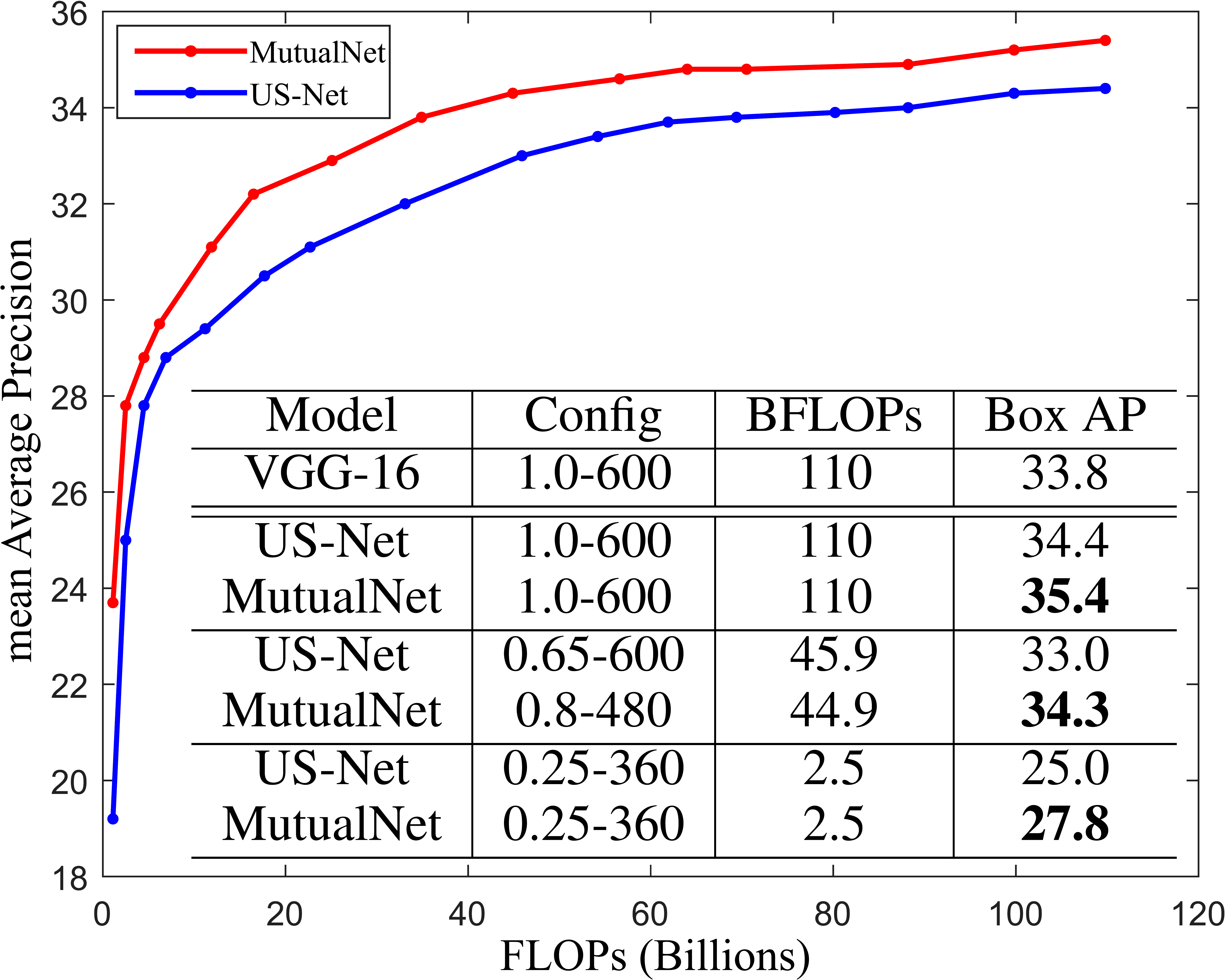} 
}
    \subfigure[Instance Segmentation]{
    \includegraphics[width=0.425\textwidth]{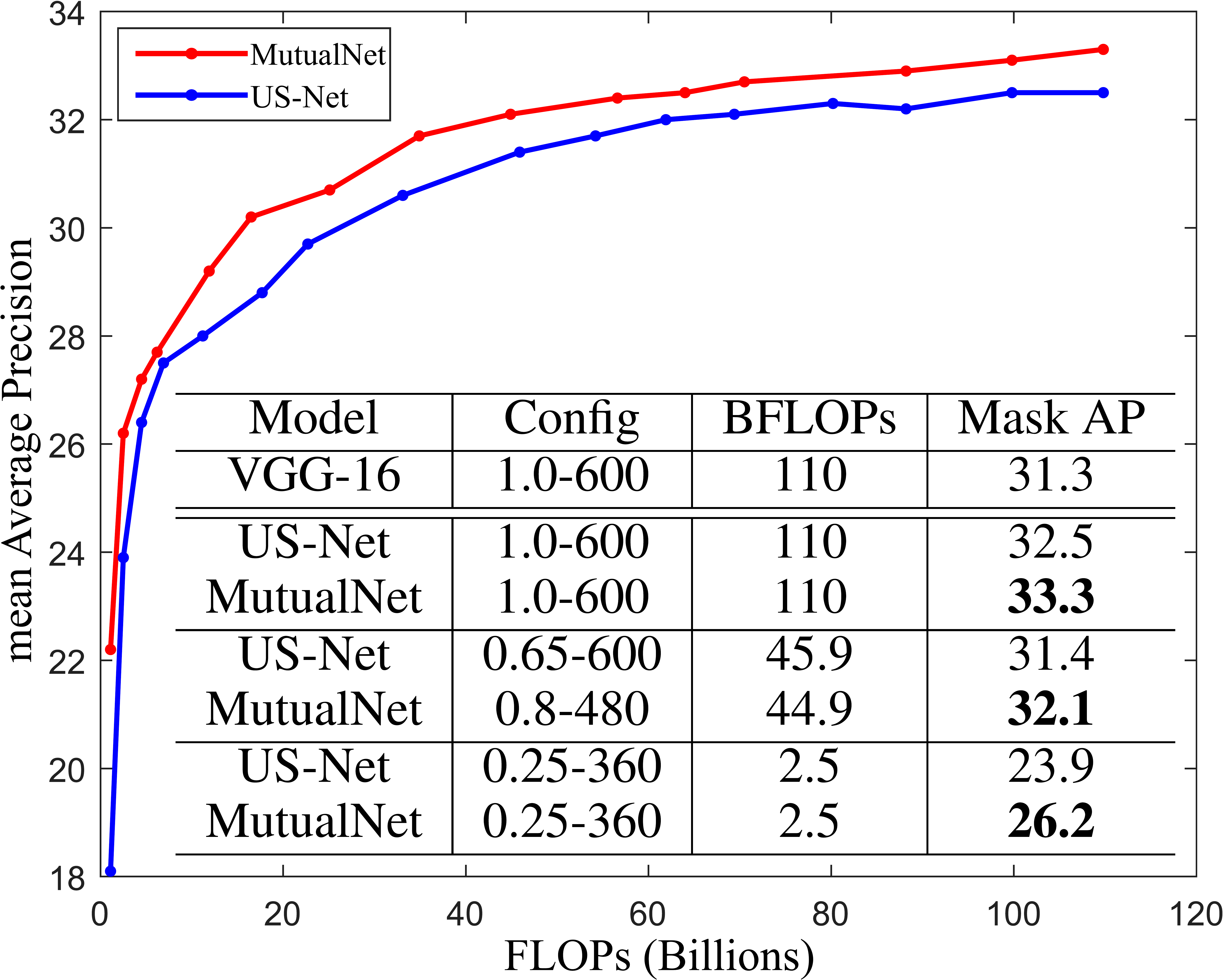}
}
    \vspace{-0.25cm}
    \caption{Average Precision - FLOPs curves of MutualNet and US-Net.}
    % (a) is bounding box average precision. (b) is mask average precision.}
    \label{det}
\end{figure}
\begin{figure}[h!]
    \centering
    \includegraphics[width=1.0\textwidth]{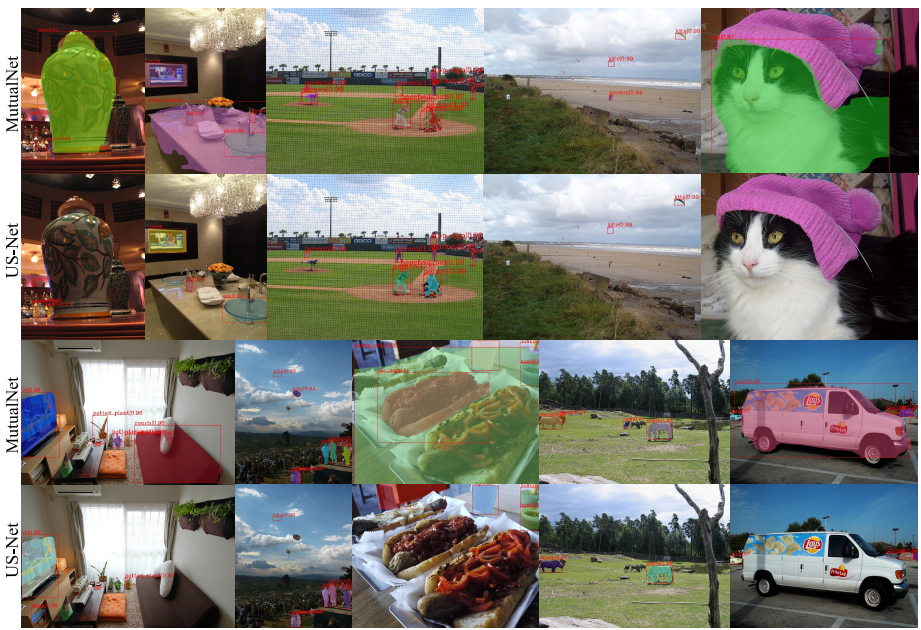}
    \caption{Object detection and instance segmentation examples.}
    \label{vis}
\end{figure}
We also evaluate our method on COCO object detection and instance segmentation \cite{coco}. The experiments are based on Mask-RCNN-FPN \cite{maskrcnn, fpn} and MMDetection \cite{chen2019mmdetection} toolbox on VGG-16 \cite{vgg} backbone. We first pre-train VGG-16 on ImageNet following US-Net and MutualNet respectively. Both methods are trained with width range [0.25, 1.0]$\times$. Then we fine-tune the pre-trained models on COCO. The FPN neck and detection head are shared among different sub-networks. For simplicity, we don't use \textit{inplace distillation}. Rather, each sub-network is trained with the ground truth. The other training procedures are the same as training ImageNet classification. Following common settings in object detection, US-Net is trained with image resolution $1000\times600$. Our method randomly selects resolutions from $1000 \times \{600, 480, 360, 240\}$. All models are trained with 2$\times$ schedule for better convergence and tested with different image resolutions. The mean Average Precision (AP at IoU=0.50:0.05:0.95) are presented in Fig. \ref{det}. These results reveal that our MutualNet significantly outperforms US-Net under all resource constraints. Specifically, for the full network (1.0$\times$-600), MutualNet significantly outperforms both US-Net and independent network. This again validates the effectiveness of our width-resolution mutual learning scheme. Fig. \ref{vis} provides some visual examples which reveal that MutualNet is more robust to small-scale and large-scale objects than US-Net.

\section{Conclusion and Future Work}
This paper highlights the importance of simultaneously considering network width and input resolution for efficient network design. A new framework namely MutualNet is proposed to mutually learn from network width and input resolution for adaptive accuracy-efficiency trade-offs. Extensive experiments have shown that it significantly improves inference performance per FLOP on various datasets and tasks. The mutual learning scheme is also demonstrated to be an effective training strategy for boosting single network performance. The generality of the proposed framework allows it to translate well to generic problem domains. This also makes logical extensions readily available by adding other network dimensions, e.g., network depth and bit-width, to the framework. The framework can also be extended to video input and 3D neural networks, where both spatial and temporal information can be leveraged.

\bibliographystyle{splncs04}
\bibliography{egbib}
\end{document}